\documentclass[10pt,twocolumn,letterpaper]{article}

\usepackage{cvpr}              

\usepackage[dvipsnames]{xcolor}

\definecolor{cvprblue}{rgb}{0.21,0.49,0.74}
\usepackage[pagebackref,breaklinks,colorlinks,citecolor=cvprblue]{hyperref}
\usepackage{hyperref}
\def\confName{CVPR}
\def\confYear{2024}

\title{HCQA @ Ego4D EgoSchema Challenge 2024}

\author{
Haoyu Zhang$^{1\,2}$, Yuquan Xie$^{1}$, Yisen Feng$^{1}$, Zaijing Li$^{1\,2}$, Meng Liu$^{3}$, Liqiang Nie$^{1}$\\
$^1$Harbin Institute of Technology (Shenzhen) \qquad  $^2$Peng Cheng Laboratory    \\$^3$Shandong Jianzhu University\\
{\tt\small \{zhang.hy.2019, lzj14011, mengliu.sdu, nieliqiang\}@gmail.com;} \\ {\tt\small \{23S051007, 23S051028\}@stu.hit.edu.cn}
}

\begin{document}
\maketitle
\begin{abstract}
In this report, we present our champion solution for Ego4D EgoSchema Challenge in CVPR 2024.
To deeply integrate the powerful egocentric captioning model and question reasoning model, 
we propose a novel \textbf{H}ierarchical \textbf{C}omprehension scheme
for egocentric video \textbf{Q}uestion \textbf{A}nswering, named \textbf{HCQA}. It consists of three stages: Fine-grained Caption Generation, Context-driven Summarization, and Inference-guided Answering. Given a long-form video, HCQA captures local detailed visual information and global summarised visual information via Fine-grained Caption Generation and Context-driven Summarization, respectively. Then in Inference-guided Answering, HCQA utilizes this hierarchical information to reason and answer given question.  On the EgoSchema blind test set, HCQA achieves 75\% accuracy in answering over 5,000 human curated multiple-choice questions. Our code will be released at \href{https://github.com/Hyu-Zhang/HCQA}{https://github.com/Hyu-Zhang/HCQA}.
\end{abstract}

\section{Introduction}
\label{sec:intro}

The Ego4D~\cite{grauman2022ego4d} EgoSchema challenge involves choosing the correct answer from five options based on a three-minute-long egocentric video and its related question. The evaluation of this challenge is performed on the EgoSchema dataset~\cite{mangalam2024egoschema}, which consists of over 5,000 human curated multiple-choice question answer pairs, spanning over 250 hours of real video data, covering a very broad range of natural human activity and behavior.
Therefore, this challenge is particularly interesting for evaluating long-context understanding, as it benefits from long ``temporal certificate'' lengths, i.e. the minimum video duration a human needs to answer the question accurately~\cite{balavzevic2024memory}.

Existing work can be broadly categorized into two groups: 1) train a powerful question answering model for egocentric videos~\cite{papalampidi2023simple, wang2024internvideo2, reid2024gemini}. This approach tends to perform better in capturing details in specific fields. However, this is often limited by large-scale data as well as time and resources. 2) fine-tune a large language model (LLM) via prompt~\cite{zhang2023simple,wang2023lifelongmemory,wang2024videoagent}. This method opens up new solution ideas to leverage existing strong LLM for adapting downstream tasks. Although this approach may be limited by LLM itself to achieve optimal performance, it requires much less computational resources and time, as well as capitalizing on the extensive knowledge embedded in LLM. Therefore, considering the time cost, we adopt the second approach to design our solution.

In Table~\ref{tab:leaderboard}, the baseline LifelongMemory~\cite{wang2023lifelongmemory} achieves the optimal performance in the available methods by applying the effective captioning model LaViLa~\cite{zhao2023learning} and the reasoning model GPT-4\footnote{\url{https://openai.com/index/gpt-4/}.}. However, we find that this method does not establish a temporal correlation between the different captions such that GPT-4 is unable to fully understand the visual scene and complete activity reflected by the captions. At the same time, in-context learning is not introduced, which has been well demonstrated to lead LLM to better perform new tasks.

Therefore, to address the above limitations, we propose a hierarchical comprehension scheme, referred to as HCQA, which incorporates both context-driven summarization and inference-guided answering. Summarization motivates the LLM to better understand the temporal information of the scene, and in-context learning makes it easier for the LLM to understand and perform EgoSchema task. By applying this simple but effective program, we surpass other teams in this challenge and our pipeline achieves a significant improvement compared to the best baseline LifelongMemory (i.e., 68\% $\rightarrow$ 75\% on accuracy in Table~\ref{tab:leaderboard}).

\begin{figure*}[t]
  \centering
   \includegraphics[width=\linewidth]{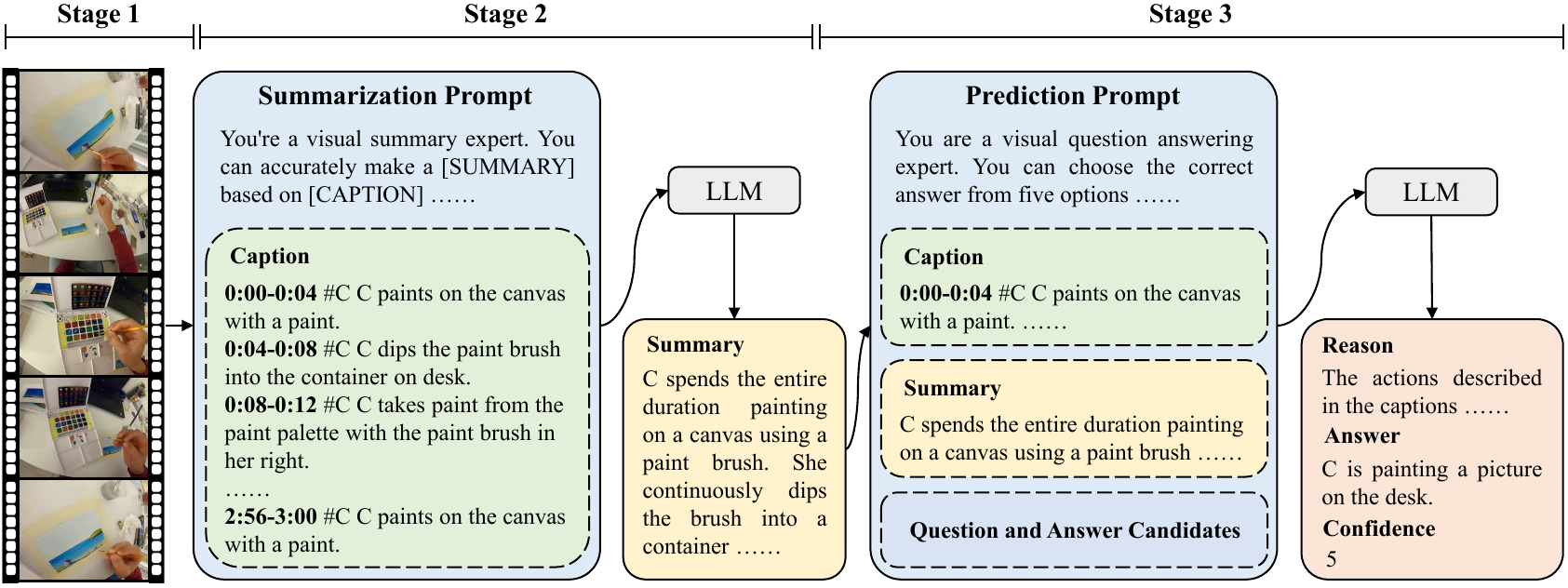}
   \caption{An illustration of our multi-stage pipeline.}
   \label{fig:method}
\end{figure*}

\section{Methodology}
\label{sec:formatting}
As shown in Figure \ref{fig:method}, 
the whole process can be described as follows: 
Given a long-form video, HCQA first transforms it into multiple local clip captions, then adopts these as input to generate a global temporal summary, and finally leverage these hierarchical descriptions to predict the answer to a given question.


\noindent\textbf{Fine-grained Caption Generation}.
To obtain detailed visual information of the video, we segment a 180-second video into clips at 4-second intervals, resulting in 45 video clips. Then, we employ LaViLa \cite{zhao2023learning} as our captioning model, which is a excellent vision-language model pretrained on Ego4D \cite{grauman2022ego4d}. Specifically, we uniformly sample 4 frames from each clip and use LaViLa to produce five corresponding textual captions for diversity. In this way, we can obtain 45$\times$5 captions, which could provide sufficient visual semantics for the subsequent question answering.

\noindent\textbf{Context-driven Summarization}.
The original caption provides a detailed description of each clip in the video, however these descriptions are discrete and unrelated. In order to integrate these captions from the temporal dimension, we propose in-context aware caption summarization, which couples these captions to establish associations and aggregates them into an overall overview of the video. Specifically, we employ prompt learning to instruct GPT-4o\footnote{\url{https://openai.com/index/hello-gpt-4o/}.} to generate a comprehensive overview of a video based on given captions. To achieve a more accurate and detailed video summary, we employ in-context learning \cite{brown2020language,dai2022can} in our approach. Empirically, we set the number of cases for in-context learning to 1. By leveraging this high-quality case as references, we guide the model to consider both local and global information during the summary generation process.

\noindent\textbf{Inference-guided Answering}.
To enhance the model's reasoning ability, we propose using the Chain-of-Thought (CoT) method \cite{wei2022chain}. This method guides the model to capture key information from captions and summaries, explicitly outputting the reasoning process for questions. Consequently, it improves the model's accuracy in answering complex visual questions. 
Notably we also employ in-context learning, using three high-quality examples from the EgoSchema subset.
Moreover, inspired by the impact of reflection mechanism \cite{shinn2023reflexion} on enhancing the performance of LLM, we incorporate a reflection mechanism into the question answering process. Specifically, after generating an answer, we prompt the model to output a confidence score for its response. If the confidence is below a certain threshold (5 in our settings), we require the model to reflect on its previous answer, assess any potential errors, and correct it if necessary. 


\section{Experiment}

\subsection{Performance Comparison}
Table~\ref{tab:leaderboard} displays the existing methods as well as our primary leaderboard results. From the results, we can see that the optimal existing method LifelongMemory achieves 68\% accuracy on the EgoSchema full set, and yet this approach can only be ranked 5th on the public leaderboard. 
Our framework achieves a accuracy of 75\%, ranked first, significantly outperforming all the other teams and existing work. This thereby proves the superiority of our method.

\subsection{Solution Evolution}

Table~\ref{tab:ablation} presents the iterative process of our solution. We first utilize the LifelongMemory~\cite{wang2023lifelongmemory} as our backbone and test its results on the EgoSchema dataset. To enhance the expression of video clip, we increase the number of caption from 1 to 3, achieving an absolute improvement of 2.4\%. After observing this obvious leap, we further increase the number of caption while introducing an example for in-context learning. And in order to strengthen the understanding of the LLM for global temporal information, we add inductive summarization in addition to predicting answers. This produces a gain of 3.2\%. In order to reduce the difficulty of performing simultaneous summarization and prediction for LLM, we separate the process into two phases: summarization followed by prediction. Although the length of the generated summaries increased significantly, this does not result in a significant accuracy improvement. Considering that one example may not provide guidance for different samples, we enrich the original example by filtering three typical examples from the subset, which gains 0.5\% increments. It is worth noting that all of the above solutions generate reason and confidence in addition to predicting answer. For uncertain predictions, i.e., low confidence, we ask LLM to reflect on and re-predict previous answers. This results in a 0.2\% benefit.

\begin{table}
  \caption{Performance comparison of existing work and the top five teams on the public leaderboard.}
  \centering
  \begin{tabular}{ccc}
    \toprule
    Method & Rank & Accuracy \\
    \midrule
    mPLUG-Owl~\cite{ye2023mplug} & -&0.31\\
    LongViViT~\cite{papalampidi2023simple}& -&0.33\\
    InternVideo2~\cite{wang2024internvideo2} &-& 0.41\\
    LLoVi~\cite{zhang2023simple} &-& 0.50\\
    VideoAgent~\cite{wang2024videoagent} & -&0.54\\
    ProViQ~\cite{choudhury2023zero}&-&0.57\\
    Gemini 1.5 Pro~\cite{reid2024gemini} & -&0.63\\
    LifelongMemory~\cite{wang2023lifelongmemory}&-&0.68\\
    \midrule
    Host\_82934\_Team & 5 & 0.64\\
    VeryLongVQA & 4 & 0.69\\
    PaMsEgoAI & 3& 0.71\\
    GMMV & 2 & 0.74\\
    \midrule
   HCQA (iLearn) & 1& \textbf{0.75}\\
    \bottomrule
  \end{tabular}
  \label{tab:leaderboard}
\end{table}

\subsection{Ablation Study}
In Table~\ref{tab:captioner}, we investigate the effects of different captioning and reasoning models. The results show that LaViLa achieves the best results over EgoVLP~\cite{lin2022egocentric} and VideoRecap~\cite{islam2024video} under consistent application of GPT-3.5. After fixing LaViLa for the captioning model, we also study the effectiveness of different reasoning models. These results suggest that GPT-4o's reasoning is superior to that of GPT-3.5 and GPT-4, leading to a superior performance.

\subsection{Case Analysis}
\begin{figure}
  \centering
  \begin{subfigure}{\linewidth}
    \includegraphics[width=\linewidth]{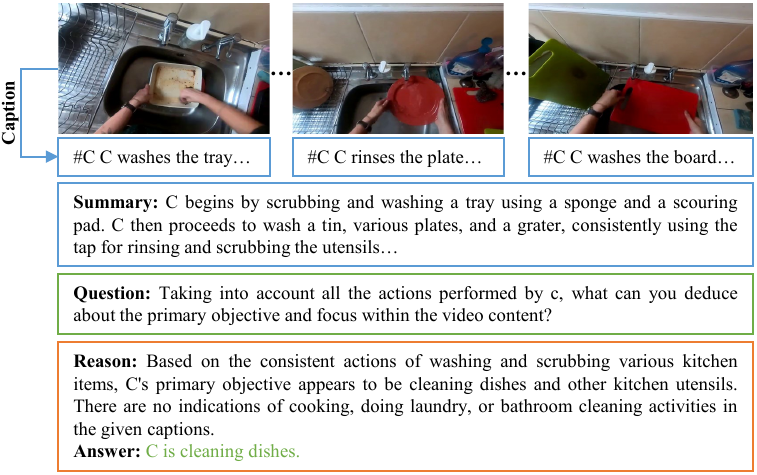}
    \caption{Success case.}
    \label{fig:success_case}
  \end{subfigure}
  \hfill
  \begin{subfigure}{\linewidth}
    \includegraphics[width=\linewidth]{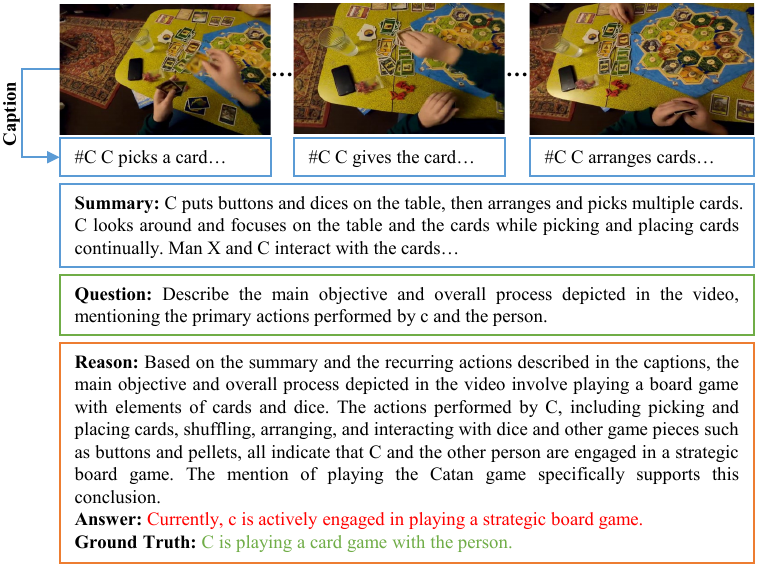}
    \caption{Failure case.}
    \label{fig:failure_case}
  \end{subfigure}
  \caption{Two examples of our framework on EgoSchema subset.}
  \label{fig:case}
  \vspace{-1em}
\end{figure}

Figure~\ref{fig:case} shows two examples of our framework, including a successful one and a failed one. In Figure~\ref{fig:success_case}, captions and summary are correctly generated to describe the camera wearer's short-term actions and long-term activities, i.e., cleaning dishes and others. With the hierarchical narrations, our framework understands the complete activity and correctly answers the question with appropriate explanations. In Figure~\ref{fig:failure_case}, although the video content is appropriately described, our reasoning model does not predict the correct answer. This may be due to the fact that the words ``card game" and ``board game" in the answer are too close to each other, causing the LLM to fail to specify the boundary between the two.


\begin{table*}
  \caption{Performance comparison on EgoSchema full set between our different solutions.}
  \centering
  \begin{tabular}{cccc}
    \toprule
    Order & Accuracy & Caption Number & Modification \\
    \midrule
    1 & 0.684 &1& Follow LifelongMemory~\cite{wang2023lifelongmemory} pipeline\\
    2 & 0.708 (0.024$\uparrow$) &3& Replace one caption with three per clip\\
    3 & 0.740 (0.032$\uparrow$) &5& Adopt one-shot in-context learning and reason both summary and answer\\
    4 & 0.741 (0.001$\uparrow$) &5&Reason the summary first and then predict the answer\\
    5 & 0.746 (0.005$\uparrow$) &5&Adopt three-shot in-context learning\\
    6 & 0.748 (0.002$\uparrow$) &5&Perform reflection for low-confidence sample\\
    \bottomrule
  \end{tabular}
  \label{tab:ablation}
  \vspace{-1ex}
\end{table*}

\begin{table}
  \caption{Accuracy of our framework on EgoSchema subset with different captioning and reasoning model.}
  \centering
  \begin{tabular}{ccccc}
    \toprule
    Captioning Model & Reasoning Model & Accuracy \\
    \midrule
    EgoVLP~\cite{lin2022egocentric}& GPT-3.5 & 0.466\\
    LaViLa~\cite{zhao2023learning} & GPT-3.5&\textbf{0.518}\\
    VideoRecap~\cite{islam2024video} & GPT-3.5&0.512\\
    \midrule
    LaViLa~\cite{zhao2023learning} & GPT-4 & 0.583\\
    LaViLa~\cite{zhao2023learning} & GPT-4o & \textbf{0.588}\\
    \bottomrule
  \end{tabular}
  \label{tab:captioner}
  \vspace{-1ex}
\end{table}

\section{Conclusion}
We present our champion solution to the Ego4D EgoSchema challenge in CVPR 2024. Experimental results have demonstrated the superiority of our scheme over all baselines. This shows that temporal summarization and multi-instance in-context learning can help LLM to better understand complex tasks. Besides, we illustrate the limitations of our pipeline through a failure case, hoping to help inspire new ideas.
{
    \small
    \bibliographystyle{ieeenat_fullname}
    \bibliography{main}

\begin{thebibliography}{27}
\providecommand{\natexlab}[1]{#1}
\providecommand{\url}[1]{\texttt{#1}}
\expandafter\ifx\csname urlstyle\endcsname\relax
  \providecommand{\doi}[1]{doi: #1}\else
  \providecommand{\doi}{doi: \begingroup \urlstyle{rm}\Url}\fi

\bibitem[Bala{\v{z}}evi{\'c} et~al.(2024)Bala{\v{z}}evi{\'c}, Shi, Papalampidi, Chaabouni, Koppula, and H{\'e}naff]{balavzevic2024memory}
Ivana Bala{\v{z}}evi{\'c}, Yuge Shi, Pinelopi Papalampidi, Rahma Chaabouni, Skanda Koppula, and Olivier~J H{\'e}naff.
\newblock Memory consolidation enables long-context video understanding.
\newblock \emph{arXiv preprint arXiv:2402.05861}, 2024.

\bibitem[Brown et~al.(2020)Brown, Mann, Ryder, Subbiah, Kaplan, Dhariwal, Neelakantan, Shyam, Sastry, Askell, et~al.]{brown2020language}
Tom Brown, Benjamin Mann, Nick Ryder, Melanie Subbiah, Jared~D Kaplan, Prafulla Dhariwal, Arvind Neelakantan, Pranav Shyam, Girish Sastry, Amanda Askell, et~al.
\newblock Language models are few-shot learners.
\newblock \emph{Advances in neural information processing systems}, 33:\penalty0 1877--1901, 2020.

\bibitem[Choudhury et~al.(2023)Choudhury, Niinuma, Kitani, and Jeni]{choudhury2023zero}
Rohan Choudhury, Koichiro Niinuma, Kris~M Kitani, and L{\'a}szl{\'o}~A Jeni.
\newblock Zero-shot video question answering with procedural programs.
\newblock \emph{arXiv preprint arXiv:2312.00937}, 2023.

\bibitem[Dai et~al.(2022)Dai, Sun, Dong, Hao, Ma, Sui, and Wei]{dai2022can}
Damai Dai, Yutao Sun, Li Dong, Yaru Hao, Shuming Ma, Zhifang Sui, and Furu Wei.
\newblock Why can gpt learn in-context? language models implicitly perform gradient descent as meta-optimizers.
\newblock \emph{arXiv preprint arXiv:2212.10559}, 2022.

\bibitem[Grauman et~al.(2022)Grauman, Westbury, Byrne, Chavis, Furnari, Girdhar, Hamburger, Jiang, Liu, Liu, et~al.]{grauman2022ego4d}
Kristen Grauman, Andrew Westbury, Eugene Byrne, Zachary Chavis, Antonino Furnari, Rohit Girdhar, Jackson Hamburger, Hao Jiang, Miao Liu, Xingyu Liu, et~al.
\newblock Ego4d: Around the world in 3,000 hours of egocentric video.
\newblock In \emph{Proceedings of the IEEE/CVF Conference on Computer Vision and Pattern Recognition}, pages 18995--19012, 2022.

\bibitem[Guan et~al.(2022)Guan, Song, Zhang, Liu, Yeh, and Chang]{10.1145/3503161.3548020}
Weili Guan, Xuemeng Song, Haoyu Zhang, Meng Liu, Chung-Hsing Yeh, and Xiaojun Chang.
\newblock Bi-directional heterogeneous graph hashing towards efficient outfit recommendation.
\newblock In \emph{Proceedings of the 30th ACM International Conference on Multimedia}, page 268–276. Association for Computing Machinery, 2022.

\bibitem[Islam et~al.(2024)Islam, Ho, Yang, Nagarajan, Torresani, and Bertasius]{islam2024video}
Md~Mohaiminul Islam, Ngan Ho, Xitong Yang, Tushar Nagarajan, Lorenzo Torresani, and Gedas Bertasius.
\newblock Video recap: Recursive captioning of hour-long videos.
\newblock \emph{arXiv preprint arXiv:2402.13250}, 2024.

\bibitem[Li et~al.(2022)Li, Tang, Zhao, and Zhu]{li2022emocaps}
Zaijing Li, Fengxiao Tang, Ming Zhao, and Yusen Zhu.
\newblock Emocaps: Emotion capsule based model for conversational emotion recognition.
\newblock In \emph{Findings of the Association for Computational Linguistics: ACL 2022}, pages 1610--1618, 2022.

\bibitem[Li et~al.(2023)Li, Lin, Wu, Liu, Tang, Zhao, and Li]{li2023unisa}
Zaijing Li, Ting-En Lin, Yuchuan Wu, Meng Liu, Fengxiao Tang, Ming Zhao, and Yongbin Li.
\newblock Unisa: Unified generative framework for sentiment analysis.
\newblock In \emph{Proceedings of the 31st ACM International Conference on Multimedia}, pages 6132--6142, 2023.

\bibitem[Li et~al.(2024{\natexlab{a}})Li, Chen, Shao, Jiang, and Nie]{li2024enhancing}
Zaijing Li, Gongwei Chen, Rui Shao, Dongmei Jiang, and Liqiang Nie.
\newblock Enhancing the emotional generation capability of large language models via emotional chain-of-thought.
\newblock \emph{arXiv preprint arXiv:2401.06836}, 2024{\natexlab{a}}.

\bibitem[Li et~al.(2024{\natexlab{b}})Li, Xie, Shao, Chen, Jiang, and Nie]{li2024optimus}
Zaijing Li, Yuquan Xie, Rui Shao, Gongwei Chen, Dongmei Jiang, and Liqiang Nie.
\newblock Optimus-1: Hybrid multimodal memory empowered agents excel in long-horizon tasks.
\newblock In \emph{NeurIPS}, 2024{\natexlab{b}}.

\bibitem[Lin et~al.(2022)Lin, Wang, Soldan, Wray, Yan, Xu, Gao, Tu, Zhao, Kong, et~al.]{lin2022egocentric}
Kevin~Qinghong Lin, Jinpeng Wang, Mattia Soldan, Michael Wray, Rui Yan, Eric~Z Xu, Difei Gao, Rong-Cheng Tu, Wenzhe Zhao, Weijie Kong, et~al.
\newblock Egocentric video-language pretraining.
\newblock \emph{Advances in Neural Information Processing Systems}, 35:\penalty0 7575--7586, 2022.

\bibitem[Mangalam et~al.(2024)Mangalam, Akshulakov, and Malik]{mangalam2024egoschema}
Karttikeya Mangalam, Raiymbek Akshulakov, and Jitendra Malik.
\newblock Egoschema: A diagnostic benchmark for very long-form video language understanding.
\newblock \emph{Advances in Neural Information Processing Systems}, 36, 2024.

\bibitem[Papalampidi et~al.(2023)Papalampidi, Koppula, Pathak, Chiu, Heyward, Patraucean, Shen, Miech, Zisserman, and Nematzdeh]{papalampidi2023simple}
Pinelopi Papalampidi, Skanda Koppula, Shreya Pathak, Justin Chiu, Joe Heyward, Viorica Patraucean, Jiajun Shen, Antoine Miech, Andrew Zisserman, and Aida Nematzdeh.
\newblock A simple recipe for contrastively pre-training video-first encoders beyond 16 frames.
\newblock \emph{arXiv preprint arXiv:2312.07395}, 2023.

\bibitem[Reid et~al.(2024)Reid, Savinov, Teplyashin, Lepikhin, Lillicrap, Alayrac, Soricut, Lazaridou, Firat, Schrittwieser, et~al.]{reid2024gemini}
Machel Reid, Nikolay Savinov, Denis Teplyashin, Dmitry Lepikhin, Timothy Lillicrap, Jean-baptiste Alayrac, Radu Soricut, Angeliki Lazaridou, Orhan Firat, Julian Schrittwieser, et~al.
\newblock Gemini 1.5: Unlocking multimodal understanding across millions of tokens of context.
\newblock \emph{arXiv preprint arXiv:2403.05530}, 2024.

\bibitem[Shinn et~al.(2023)Shinn, Labash, and Gopinath]{shinn2023reflexion}
Noah Shinn, Beck Labash, and Ashwin Gopinath.
\newblock Reflexion: an autonomous agent with dynamic memory and self-reflection.
\newblock \emph{arXiv preprint arXiv:2303.11366}, 2023.

\bibitem[Wang et~al.(2024{\natexlab{a}})Wang, Zhang, Zohar, and Yeung-Levy]{wang2024videoagent}
Xiaohan Wang, Yuhui Zhang, Orr Zohar, and Serena Yeung-Levy.
\newblock Videoagent: Long-form video understanding with large language model as agent.
\newblock \emph{arXiv preprint arXiv:2403.10517}, 2024{\natexlab{a}}.

\bibitem[Wang et~al.(2023)Wang, Yang, and Ren]{wang2023lifelongmemory}
Ying Wang, Yanlai Yang, and Mengye Ren.
\newblock Lifelongmemory: Leveraging llms for answering queries in egocentric videos.
\newblock \emph{arXiv preprint arXiv:2312.05269}, 2023.

\bibitem[Wang et~al.(2024{\natexlab{b}})Wang, Li, Li, Yu, He, Chen, Pei, Zheng, Xu, Wang, et~al.]{wang2024internvideo2}
Yi Wang, Kunchang Li, Xinhao Li, Jiashuo Yu, Yinan He, Guo Chen, Baoqi Pei, Rongkun Zheng, Jilan Xu, Zun Wang, et~al.
\newblock Internvideo2: Scaling video foundation models for multimodal video understanding.
\newblock \emph{arXiv preprint arXiv:2403.15377}, 2024{\natexlab{b}}.

\bibitem[Wei et~al.(2022)Wei, Wang, Schuurmans, Bosma, Xia, Chi, Le, Zhou, et~al.]{wei2022chain}
Jason Wei, Xuezhi Wang, Dale Schuurmans, Maarten Bosma, Fei Xia, Ed Chi, Quoc~V Le, Denny Zhou, et~al.
\newblock Chain-of-thought prompting elicits reasoning in large language models.
\newblock \emph{Advances in Neural Information Processing Systems}, 35:\penalty0 24824--24837, 2022.

\bibitem[Ye et~al.(2023)Ye, Xu, Xu, Ye, Yan, Zhou, Wang, Hu, Shi, Shi, et~al.]{ye2023mplug}
Qinghao Ye, Haiyang Xu, Guohai Xu, Jiabo Ye, Ming Yan, Yiyang Zhou, Junyang Wang, Anwen Hu, Pengcheng Shi, Yaya Shi, et~al.
\newblock mplug-owl: Modularization empowers large language models with multimodality.
\newblock \emph{arXiv preprint arXiv:2304.14178}, 2023.

\bibitem[Zhang et~al.(2023{\natexlab{a}})Zhang, Lu, Islam, Wang, Yu, Bansal, and Bertasius]{zhang2023simple}
Ce Zhang, Taixi Lu, Md~Mohaiminul Islam, Ziyang Wang, Shoubin Yu, Mohit Bansal, and Gedas Bertasius.
\newblock A simple llm framework for long-range video question-answering.
\newblock \emph{arXiv preprint arXiv:2312.17235}, 2023{\natexlab{a}}.

\bibitem[Zhang et~al.(2021)Zhang, Liu, Gao, Lei, Wang, and Nie]{10.1145/3474085.3475234}
Haoyu Zhang, Meng Liu, Zan Gao, Xiaoqiang Lei, Yinglong Wang, and Liqiang Nie.
\newblock Multimodal dialog system: Relational graph-based context-aware question understanding.
\newblock In \emph{Proceedings of the 29th ACM International Conference on Multimedia}, page 695–703. Association for Computing Machinery, 2021.

\bibitem[Zhang et~al.(2023{\natexlab{b}})Zhang, Liu, Li, Yan, Gao, Chang, and Nie]{10239469}
Haoyu Zhang, Meng Liu, Yuhong Li, Ming Yan, Zan Gao, Xiaojun Chang, and Liqiang Nie.
\newblock Attribute-guided collaborative learning for partial person re-identification.
\newblock \emph{IEEE Transactions on Pattern Analysis and Machine Intelligence}, 45\penalty0 (12):\penalty0 14144--14160, 2023{\natexlab{b}}.

\bibitem[Zhang et~al.(2023{\natexlab{c}})Zhang, Liu, Wang, Cao, Guan, and Nie]{zhang2023uncovering}
Haoyu Zhang, Meng Liu, Yaowei Wang, Da Cao, Weili Guan, and Liqiang Nie.
\newblock Uncovering hidden connections: Iterative tracking and reasoning for video-grounded dialog.
\newblock \emph{arXiv preprint arXiv:2310.07259}, 2023{\natexlab{c}}.

\bibitem[Zhang et~al.(2024)Zhang, Liu, Liu, Song, Wang, and Nie]{pmlr-v235-zhang24aj}
Haoyu Zhang, Meng Liu, Zixin Liu, Xuemeng Song, Yaowei Wang, and Liqiang Nie.
\newblock Multi-factor adaptive vision selection for egocentric video question answering.
\newblock In \emph{Proceedings of the 41st International Conference on Machine Learning}, pages 59310--59328. PMLR, 2024.

\bibitem[Zhao et~al.(2023)Zhao, Misra, Kr{\"a}henb{\"u}hl, and Girdhar]{zhao2023learning}
Yue Zhao, Ishan Misra, Philipp Kr{\"a}henb{\"u}hl, and Rohit Girdhar.
\newblock Learning video representations from large language models.
\newblock In \emph{Proceedings of the IEEE/CVF Conference on Computer Vision and Pattern Recognition}, pages 6586--6597, 2023.

\end{thebibliography}
}


\end{document}